\documentclass{article}
\usepackage{spconf,amsmath,graphicx}
\usepackage{wrapfig}
\usepackage[ruled,vlined]{algorithm2e}
\usepackage{enumitem}
\usepackage{stfloats}
\usepackage{caption}
\usepackage{subcaption}
\usepackage{booktabs}
\usepackage{tabularx}
\usepackage{array}
\usepackage{ragged2e}
\newcolumntype{P}[1]{>{\RaggedRight\hspace{0pt}}p{#1}}

\usepackage{tabularx,booktabs}
\newcolumntype{C}{>{\centering\arraybackslash}X} 
\usepackage{lipsum}

\usepackage{color}
  \definecolor{LightGray}{gray}{0.95}
  \definecolor{MidGray}{gray}{0.85}
  \definecolor{dark-red}{rgb}{0.4,0.15,0.15}
  \definecolor{gray-red}{rgb}{1,0.75,0.75}
  \definecolor{dark-blue}{rgb}{0.15,0.15,0.33}
  \definecolor{medium-blue}{rgb}{0,0,0.5}

\usepackage{xspace}
\makeatletter
\DeclareRobustCommand\onedot{\futurelet\@let@token\@onedot}
\def\@onedot{\ifx\@let@token.\else.\null\fi\xspace}
\def\eg{e.g\onedot}

\def\etal{et al\onedot}

\usepackage{hyperref}
  \hypersetup{
    colorlinks,
    linkcolor={dark-blue},
    citecolor={dark-blue},
    urlcolor={medium-blue}
  }
\title{
PSEUDO-LABEL REFINEMENT USING SUPERPIXELS FOR SEMI-SUPERVISED BRAIN TUMOUR SEGMENTATION}

\name{Bethany H. Thompson $^{\dagger \star}$ \qquad Gaetano Di Caterina $^{\star}$ \qquad Jeremy P. Voisey $^{\dagger}$}
  
\address{$^{\dagger}$ Canon Medical Research Europe Ltd., AI Research, Edinburgh, UK \\ $^{\star}$ University of Strathclyde, Electronic \& Electrical Engineering, Glasgow, UK}

\begin{document}
\ninept

%
\maketitle
\begin{abstract}
Training neural networks using limited annotations is an important problem in the medical domain. Deep Neural Networks (DNNs) typically require large, annotated datasets to achieve acceptable performance which, in the medical domain, are especially difficult to obtain as they require significant time from expert radiologists. Semi-supervised learning aims to overcome this problem by learning segmentations with very little annotated data, whilst exploiting large amounts of unlabelled data. However, the best-known technique, which utilises inferred pseudo-labels, is vulnerable to inaccurate pseudo-labels degrading the performance. We propose a framework based on superpixels - meaningful clusters of adjacent pixels - to improve the accuracy of the pseudo labels and address this issue. Our framework combines superpixels with semi-supervised learning, refining the pseudo-labels during training using the features and edges of the superpixel maps. This method is evaluated on a multimodal magnetic resonance imaging (MRI) dataset for the task of brain tumour region segmentation. Our method demonstrates improved performance over the standard semi-supervised pseudo-labelling baseline when there is a reduced annotator burden and only 5 annotated patients are available. We report DSC=0.824 and DSC=0.707 for the test set whole tumour and tumour core regions respectively. 
\end{abstract}
\section{Introduction}
\label{sec:intro}

In medical imaging, segmentation of pathology helps clinicians to diagnose the severity of disease, make a recommendation for treatment and monitor the response to treatment over time \cite{Suetens_1993, Hosny_2018}. However, training accurate medical imaging models requires vast amounts of expertly annotated data which can be challenging, time-consuming and prone to error and user bias, with annotation taking approximately 1 hour per patient for the 2020 Brain Tumour Segmentation dataset (BraTS 2020) \cite{Brats_1}. Developing automatic segmentation tools trained with limited labelled data is a challenging task, especially in pathology segmentation where the object varies notably between patients and there is low contrast between the object and the background.

Semi-supervised learning, which utilises both labelled and unlabelled data, is a popular way to take advantage of large amounts of unlabelled data and maintain a low annotator burden. Existing pseudo-label-based methods obtain pseudo-labels by first training a segmentation model with limited labelled data and then inferring pseudo-labels for the unlabelled data before subsequent training on both \cite{Dong-HyunLee2013, Chen2020}. However, these methods may generate inaccurate pseudo-labels and degrade the subsequent training process and can be sensitive to the mask threshold with pseudo-labels being either over or under confident \cite{Li2020, Chapelle2009}.

To overcome this problem, some previous works have suggested combining active contours, a classic unsupervised segmentation technique, with semi-supervised deep learning to refine the pseudo-labels inferred during semi-supervised training \cite{Ma2020}. The general idea is that the pseudo-labels provide the seed for region growing before using the updated pseudo labels for subsequent semi-supervised training. However, traditional boundary-based methods which may be used for region growing (\eg active contours/snakes \cite{chan_vese2001}) can be time consuming, with algorithms taking many iterations before convergence \cite{book1}, especially in 3D. To overcome this problem, region growing on the superpixel level rather than on the pixel level is proposed. 

Superpixels are the result of iterative grouping of pixels into an irregular grid of regions that mould well to salient features and edges within the image \cite{Achanta2012}. The most popular algorithm for generating superpixels is the Simple Linear Iterative Clustering (SLIC) algorithm \cite{Achanta2012}, which adapts a k-means clustering approach to generate superpixels. Although originally developed for natural images, SLIC has been shown to be effective in MRI \cite{Tian2015}, CT \cite{Qin2018} and Ultrasound \cite{Daoud2019} and is therefore able to detect the, sometimes subtle, region boundaries found in medical images. Previous work done by Borovec \etal (2017) \cite{Borovec2017} developed a superpixel region growing algorithm with a learned shape prior for segmenting individual eggs in microscopy images of Drosophila ovaries. However, in the case of pathology, such as brain tumour segmentation, a circular shape prior is not suitable as there is a large variance in topology and extent of disease between patients. Chaibou \etal (2018) \cite{Chaibou2018} developed a strategy for unsupervised segmentation of natural images based on superpixel clustering. Superpixels to be merged are assumed to satisfy two important criteria: spatial adjacency and perceptual similarity. This suggests that an efficient approach should be able to pick the most visually similar spatial neighbour of the region of interest. They argue that taking advantage of the initial clustering performed by SLIC before further clustering of those superpixels into object regions allows more efficient semantic feature extraction compared to their pixel-level counterparts.

Another related work was presented by Wang \etal (2021) \cite{Wang2021}, where a filtering mechanism was developed to remove the pseudo-labels with the lowest confidence. The work presented in our paper takes a different approach and refines the predicted pseudo-labels, making them more robust rather than reducing the dataset size through filtering. 

\begin{figure}[tp]
\centering
\centerline{\includegraphics[width=8.5cm]{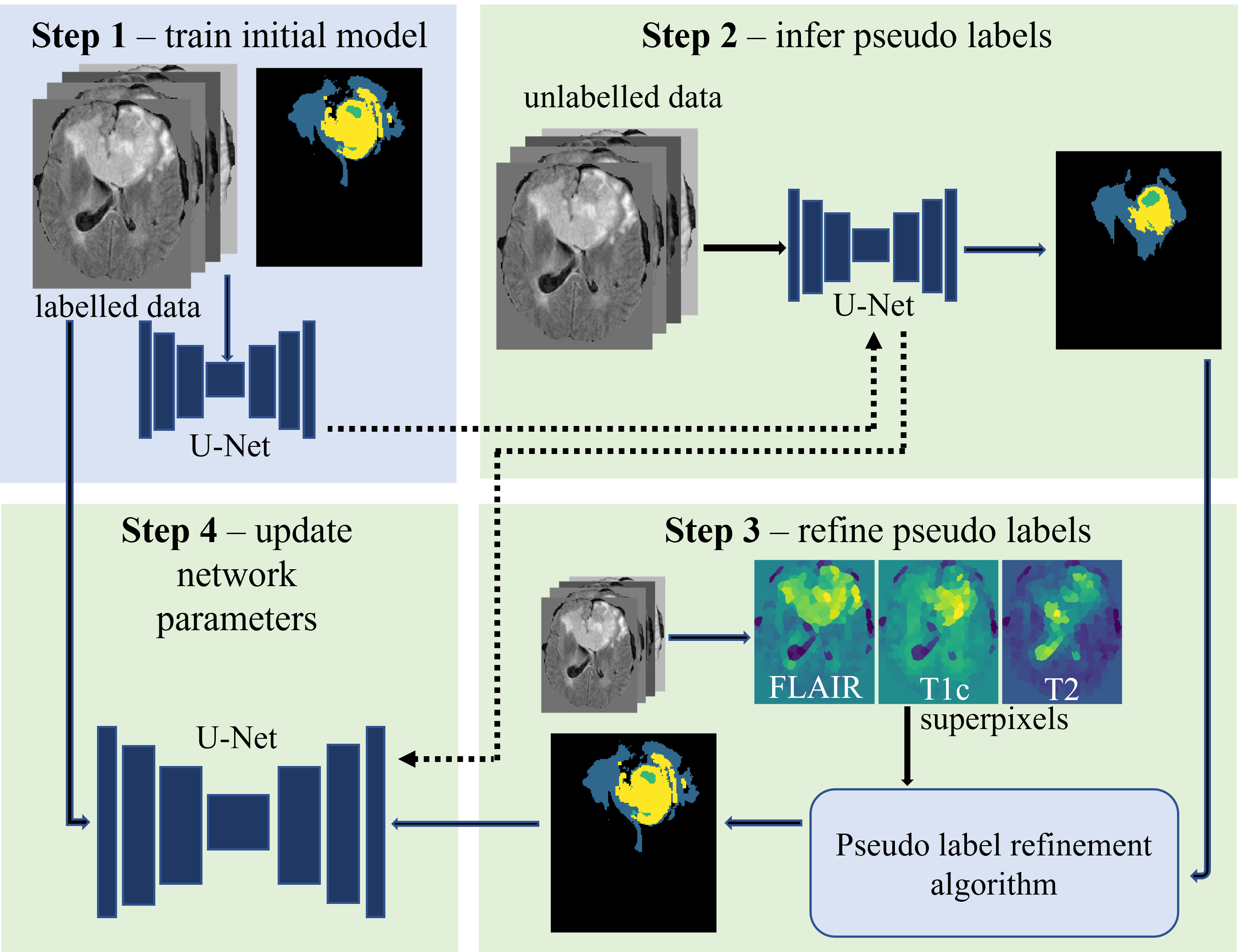}}
  \caption{Semi-supervised pseudo-labelling segmentation pipeline with the proposed superpixel-based pseudo-label refinement step}
  \label{fig:pipeline}
\end{figure}

This paper presents a modified version of the pseudo label-based semi-supervised framework combined with a superpixel-based region growing algorithm - inspired by the work of Chaibou \etal (2018) - to refine the network predicted pseudo labels and aid subsequent semi-supervised training. The proposed method is demonstrated for the task of brain tumour segmentation in multimodal MRI. \autoref{fig:pipeline} shows the proposed 4-step pipeline which incorporates superpixel region refinement after the pseudo-label update step. 

The contributions of this work are summarised as follows:
\begin{enumerate}[nolistsep]
    \item A simple 3D in-training algorithm to refine network-inferred pseudo-labels using superpixel information
    \item A similarity measure for sub-region refinement based on comparison of different MRI channels (\autoref{fig:mri_modalities}) which mimics how expert annotators determine region boundaries
    \item Demonstration of pseudo-label refinement for brain tumour segmentation which have varying topology and low contrast
\end{enumerate}

\section{METHOD}

\subsection{Semi-supervised with Pseudo-labels}
\label{sec:pseudo_label}

We implement a modified version of the original pseudo-label method proposed by Lee (2013) \cite{Dong-HyunLee2013} as follows:
\begin{enumerate}[nolistsep]
    \item The network is trained on $p_1$ ground truth (GT) labelled patients for 200 epochs
    \item This model is then used to infer the pseudo labels for the $p_2$ unlabelled patients
    \item The network is then trained in a supervised manner on both GT labelled patients and pseudo-labelled patients for the remaining 800 epochs 
\end{enumerate}

 For unlabelled data, pseudo-labels are only re-calculated every 200 epochs to avoid long training times. As training progresses, more pseudo-labelled batches are introduced into the training, according to a dynamic weighting factor $\alpha(e)$ as described by (\ref{eqn:pseudolabel_proportion}) \cite{Dong-HyunLee2013}. The number of pseudo-labelled batches per epoch as training progresses is described by (\ref{eq:2}), where $N_{e}$ is the number of pseudo-labelled batches in epoch e and $N_{T}$ is the total number of batches in an epoch.  
 
\begin{equation}
      \alpha(e) =
      \begin{cases}
      0 & e < T_1 \\
      \frac{e - T_1}{T_2 - T_1}\alpha_{f}  & T_1 \leq e  < T_2  \\
      \alpha_{f} & e  \geq T_2
      \end{cases}
\label{eqn:pseudolabel_proportion}
\end{equation}

\begin{equation}
\small
    N_{e} =  N_{T}  \frac{\alpha(e)}{\alpha(e) + 1}
\label{eq:2}
\end{equation}

\subsection{Superpixels}
\label{ssec:subhead}

Using superpixels allows feature statistics to be measured on a naturally adaptive domain rather than on a fixed window. Since superpixels tend to preserve boundaries, there is an opportunity to create an accurate segmentation by simply finding the superpixels which are part of the tumour region.

\begin{figure}[t]
    \begin{subfigure}{0.12\textwidth}
        \includegraphics[width=\linewidth]{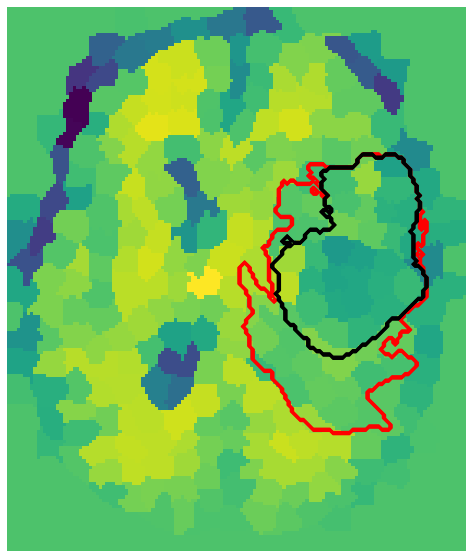}
        \caption{}
        \label{fig:region_growing_a}
    \end{subfigure}%
    \begin{subfigure}{0.12\textwidth}
        \includegraphics[width=\linewidth]{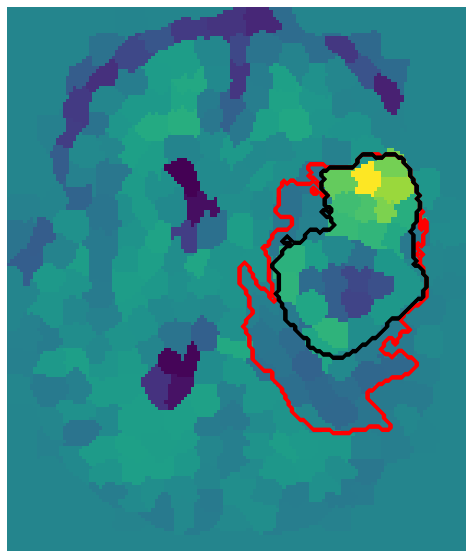}
        \caption{}
        \label{fig:region_growing_b}
    \end{subfigure}%
    \begin{subfigure}{0.12\textwidth}
        \includegraphics[width=\linewidth]{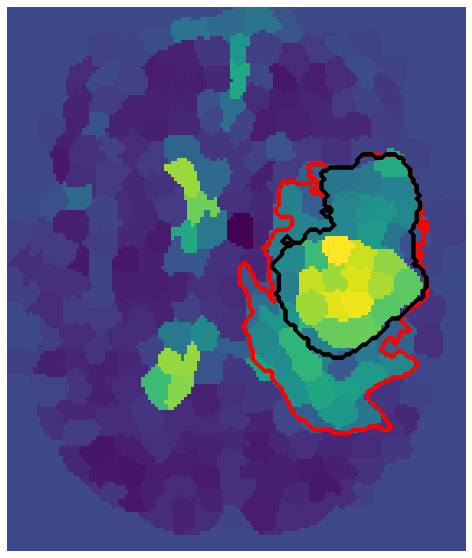}
        \caption{}
        \label{fig:region_growing_c}
    \end{subfigure}%
    \begin{subfigure}{0.12\textwidth}
        \includegraphics[width=\linewidth]{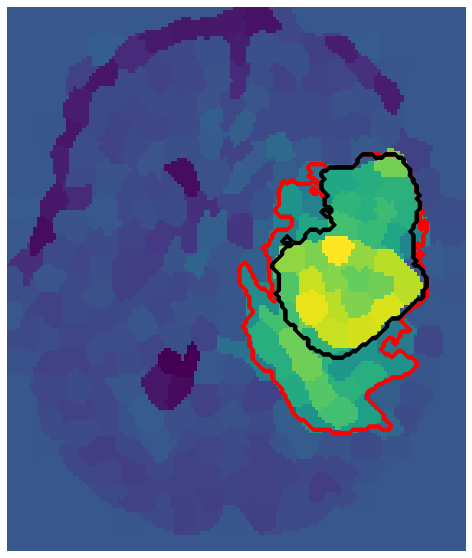}
        \caption{}
        \label{fig:region_growing_d}
    \end{subfigure}
    \caption{Superpixel maps with sigma=1, compactness=0.01, $n_{segments}$ = 350 for patient 224. Red contour: WT GT boundary, Black contour: TC GT boundary (a) T1, (b) T1Gd, (c) T2 and (d) T2-FLAIR}
    \label{fig:superpixels}
\end{figure}

The superpixel masks were generated for each unlabelled training volume using the SLIC algorithm \cite{Achanta2012}. \autoref{fig:superpixels} shows the superpixel maps with sigma=1, compactness=0.01, $n_{segments}$ = 350 for the four different MRI modalities used in this work. The overlaid label boundaries for the two regions of interest - Whole tumour (WT) and Tumour Core (TC) - are shown in red and black respectively. The annotator protocol used for the original BraTS dataset \cite{Brats_1, Brats_2, Brats_3} states that different regions are more easily distinguishable by using different MRI modalities as shown in both \autoref{fig:superpixels} and \autoref{fig:mri_modalities}. To mimic this, to refine the WT region, a 3D superpixel map generated from the T2-FLAIR scan should be used. Similarly, if interested in refining pseudo-labels for the tumour core region then the superpixel maps generated from the T2 and T1Gd scans should be used. 

\begin{algorithm}[htb]
\small
\SetAlgoLined
\KwData{$Image$, $sim_0$, $p_0$, $n_c$} 
\KwResult{$p_r$}
 $s_p \leftarrow \text{Slic}(Image)$\\
 $p_r \leftarrow \text{FitPseudolabelToSuperPixels}(s_p, p_0)$\\
 
 \Repeat{$merged = false$}{
 $N_p \leftarrow \text{FindNeighbours}(s_p, p_r)$\\
      $merged \leftarrow false$\\
      
      $count \leftarrow 0$\\
      \Repeat{$merged$ \textbf{or} $count > n_c$}{
      $count \leftarrow count + 1$\\
      $sim_p \leftarrow \{ \text{Sim}(p_r, n) \forall n : n\in N_p$ \} \\
      $n_c^* \leftarrow \text{argmax}(sim_p)$ \\
        $N_c \leftarrow \text{FindNeighbours}(s_p, n_c^*)$\\
      $sim_c \leftarrow \{ \text{Sim}(n_c^*, n) \forall n : n\in N_c \}$ \\
      
      $n_q^* \leftarrow \text{argmax}(sim_c) $\\
      
        \If{$n_q^*$ \textbf{in} $p_r$}{
            \If{$\text{Sim}(p_r,n_c^*)>sim_0$}{
                $merged \leftarrow true$\\
                $p_r \leftarrow \text{Merge}(p_r, n_c^*$) \\
                
                }}
            $N_p \leftarrow N_p \setminus n_c^*$
        }
        }
\caption{Pseudo-label seeded superpixel region refinement}
\label{alg:region_growing}
\end{algorithm}

\subsection{Pseudo-Label Refinement}
\label{sssec:PLRefine}

The proposed pipeline consists of the semi-supervised pseudo-labelling method described in \autoref{sec:pseudo_label}, with the addition of a pseudo-label seeded region refinement algorithm as shown in \autoref{fig:pipeline}.

A feature vector was calculated for each superpixel which comprises 9 intensity, texture and gradient based statistical features (mean, variance, skewness, 10-bin intensity histogram, contrast, energy, entropy, 10-bin histogram of gradient orientation and 10-bin histogram of gradient magnitude). Inspired by Chaibou \etal (2018) \cite{Chaibou2018}, the proposed merging procedure is  based on the agglomerative clustering algorithm via the Ward method \cite{ward1963}. The superpixel similarity measure is defined in the same way as Chaibou \etal (2018), where both content similarity and border similarity make up the overall similarity measure. The important difference in our work is that the region refinement algorithm is initially seeded with the network-inferred pseudo-labels as a starting point, where clustering is constrained to the pseudo-label region rather than general unsupervised clustering throughout the image. Furthermore, our work is in 3D rather than 2D and utilises 3-channels of low contrast MRI images rather than natural images. The proposed superpixel region refinement algorithm is shown in \autoref{alg:region_growing}. As the pseudo-label region grows, the similarity between the region and its neighbouring superpixels is recalculated after each merging operation. This is in recognition that as the region grows, the similarity between the region and its neighbours will also change. The refinement of a pseudo-label stops when there are no more superpixel neighbours which satisfy the merging conditions: the similarity must be above the stopping similarity $sim_0$ and the region neighbour must mutually choose the pseudo-label region as its most similar neighbour.

The variables in \autoref{alg:region_growing} are defined as follows: $Image$: 3D MRI patient scan, $sim_0$: user-defined stopping similarity, $p_0$: initial 3D pseudo-label predicted by the network, $n_c$: hyperparameter which defines the number of candidate neighbours to check for mutual similarity with the pseudo-label region in a given iteration, $s_p$: superpixel map, $p_r$: refined pseudo-label, $N_p$: the superpixel neighbours of the pseudo-label region, $sim_p$: similarities between a merging candidate and the pseudo-label region, $n_c^*$: the candidate with the highest similarity to the pseudo-label region, $N_c$: the superpixel neighbours of a given merging candidate, $sim_c$: similarities between a merging candidate and each of its neighbours, $n_q^*$: the neighbour q with the highest similarity to a given merging candidate.

\section{EXPERIMENT SETUP}

\subsection{Dataset}
\label{sec:data}

The data used in this paper is from the BraTS 2020 dataset \cite{Brats_1, Brats_2, Brats_3}, comprising 369 patients of multimodal MRI scans of glioblastoma (GBM) and lower grade glioma (LGG) with accompanying ground truth labels.

\begin{figure}[t]
    \begin{subfigure}{0.12\textwidth}
        \includegraphics[width=\linewidth]{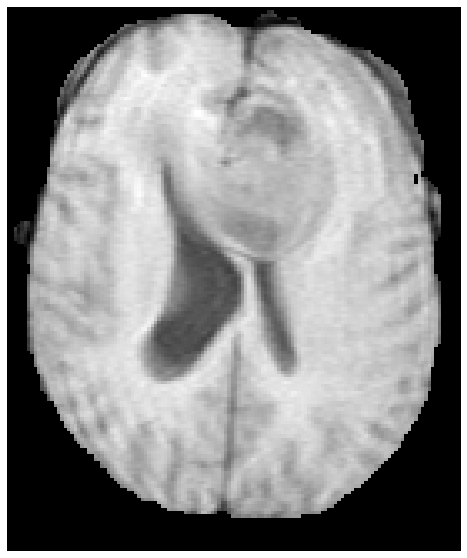}
        \caption{}
        \label{fig:mri_t1}
    \end{subfigure}%
    \begin{subfigure}{0.12\textwidth}
        \includegraphics[width=\linewidth]{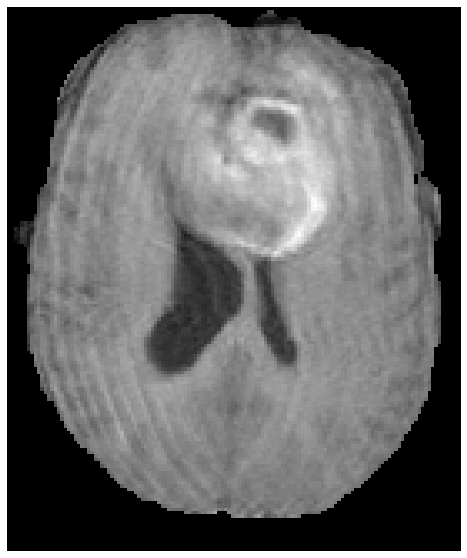}
        \caption{}
        \label{fig:mri_t1ce}
    \end{subfigure}%
    \begin{subfigure}{0.12\textwidth}
        \includegraphics[width=\linewidth]{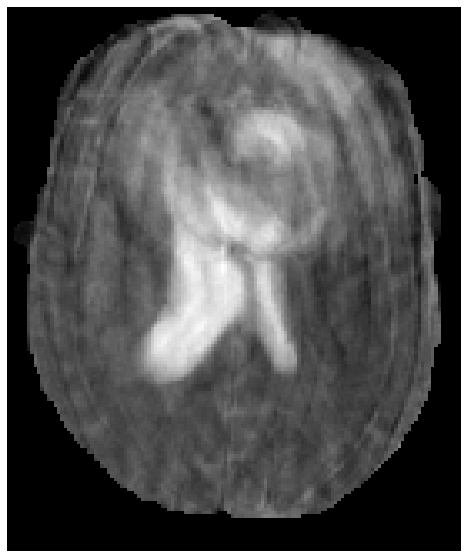}
        \caption{}
        \label{fig:mri_t2}
    \end{subfigure}%
    \begin{subfigure}{0.12\textwidth}
        \includegraphics[width=\linewidth]{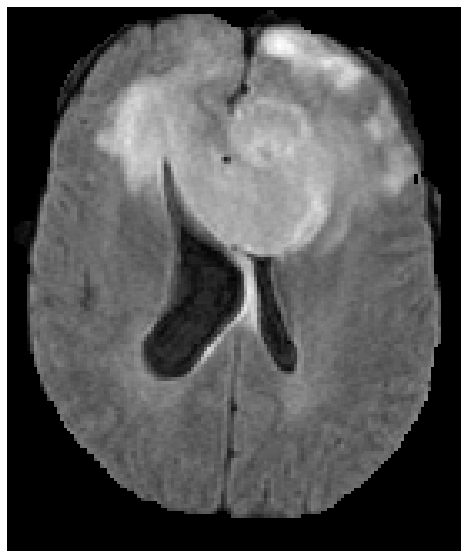}
        \caption{}
        \label{fig:mri_flair}
    \end{subfigure}

    \begin{subfigure}{0.12\textwidth}
        \includegraphics[width=\linewidth]{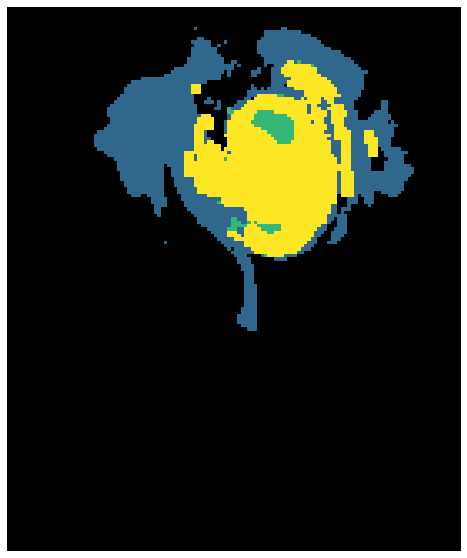}
        \caption{}
        \label{fig:brats_annotations_all}
    \end{subfigure}%
    \begin{subfigure}{0.12\textwidth}
        \includegraphics[width=\linewidth]{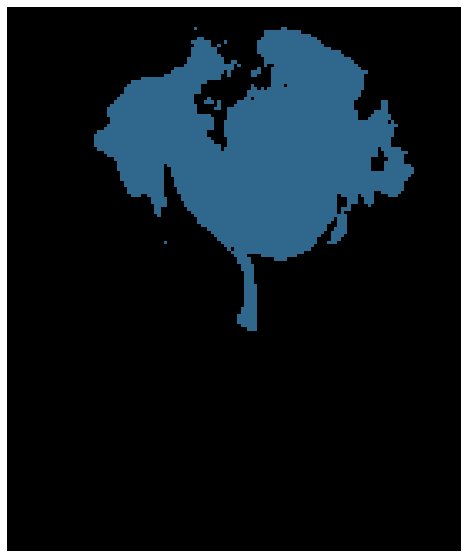}
        \caption{}
        \label{fig:brats_annotations_whole}
    \end{subfigure}%
    \begin{subfigure}{0.12\textwidth}
        \includegraphics[width=\linewidth]{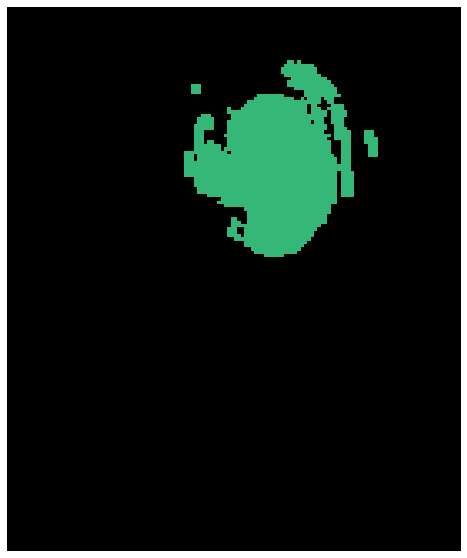}
        \caption{}
        \label{fig:brats_annotations_core}
    \end{subfigure}%
    \begin{subfigure}{0.12\textwidth}
        \includegraphics[width=\linewidth]{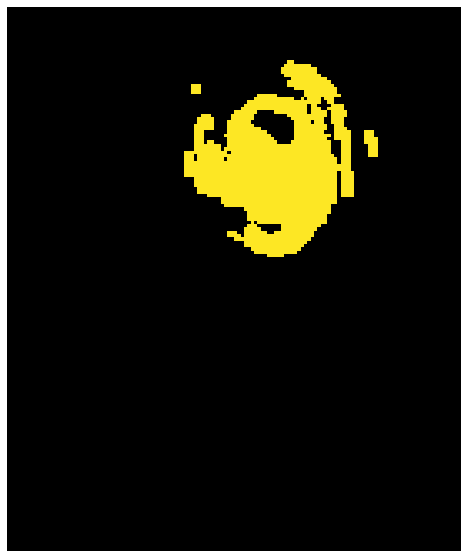}
        \caption{}
        \label{fig:brats_annotations_enhancing}
    \end{subfigure}

    \caption{The different MRI modalities present in BraTS. (a) native (T1), (b) post-contrast T1-weighted (T1Gd), (c) T2-weighted (T2), (d) T2 Fluid Attenuated Inversion Recovery (T2-FLAIR), (e) original BraTS annotations, (f)-(h) region-based annotations: whole tumour (blue), tumour core (green), enhancing tumour (yellow)}
    \label{fig:mri_modalities}

\end{figure}

The data was split between training and testing with a ratio of 9:1, resulting in a training set of 331 patients and a hold-out test set of 38 patients. The data contains four different MRI modalities (native (T1), post-contrast T1-weighted (T1Gd), T2-weighted (T2), and T2 Fluid Attenuated Inversion Recovery (T2-FLAIR)) for each patient as shown in \autoref{fig:mri_modalities}. It should be noted that the visibility of the different tumour regions depends on the MRI modality used. 

\autoref{fig:brats_annotations_all} shows the original tumour regions annotated in the BraTS 2020 dataset: Background, Edema, Active Enhancing and Non-enhancing. \autoref{fig:brats_annotations_whole} and \ref{fig:brats_annotations_core} show the regions of interest for this work: whole tumour (WT) and tumour core (TC) respectively.

\subsection{Implementation Details}
\subsubsection{U-Net Training Framework}
\label{sec:backbone}

The nnU-Net framework \cite{nnUnet_1, nnUnet_2} was used in this work and its trainer was modified for semi-supervised training. The U-Net makes a prediction for the WT and TC regions for each pixel.
The default training protocols are used: an initial learning rate of 0.01 updated by the ‘polyLR’ learning rate schedule \cite{chen2017}, stochastic gradient descent optimiser with momentum (0.99), all models were trained with a fixed length of 1000 epochs and batch size 2. Patch-based training was performed with a patch size of 128$\times$128$\times$128 and training was performed for 125,000 iterations. 

Data augmentation, as in \cite{nnUnet_1, nnUnet_2},  was applied during training and patch oversampling of patches containing at least one of the foreground classes was implemented, so that 33.3\% of patches were guaranteed to contain one of the foreground classes present in the selected training sample. The loss function was an equally-weighted combination of binary cross-entropy and dice loss. 

\subsubsection{Pseudo-label Baseline}
\label{sec:pseudo_label_implementation}

The pseudo-labelling method is described in \autoref{sec:pseudo_label} where the number of labelled patients is 5 and the number of unlabelled patients is 259. As training progresses, more pseudo-labelled batches are introduced into the training as described by the dynamic weighting factor $\alpha(e)$ in (\ref{eqn:pseudolabel_proportion}). The hyperparameters were set to be $\alpha_{f}$ = 3, $T_1$ = 200, $T_2$ = 700. The total number of batches in an epoch, $N_{T}$, is defined as 250 in this work. The hyperparameter $N_{e}$, the number of pseudo-labelled batches in epoch e as defined in (\ref{eq:2}), translates to there being 2 pseudo-labelled batches at epoch 200 (1\% of the epoch) to 188 unlabelled batches from epoch 700 onwards (75\% of the epoch).

\subsubsection{Proposed Pipeline}
\label{sssec:subsubhead}

The proposed pipeline is shown in \autoref{fig:pipeline}, which incorporates superpixel region refinement after the pseudo-label update step. In step 1, the network is trained in a fully-supervised manner on only the labelled data. In step 2, the trained network is used to infer pseudo-labels for the unlabelled data. In step 3, the pseudo-labels are refined by the superpixel region refinement algorithm as described in \autoref{sssec:PLRefine}. In step 4, the network is trained on both the labelled and pseudo-labelled data. Steps 2-4 are repeated as described in \ref{sec:pseudo_label}. 

The U-Net training framework was a modified version of nnU-Net for semi-supervised training as described in \autoref{sec:backbone}, \autoref{sec:pseudo_label} and \autoref{sec:pseudo_label_implementation}. The region refinement algorithm shown in \autoref{alg:region_growing} was run after every pseudo-label update step (every 200 epochs). The stopping similarity was empirically set to $sim_0$=0.1 and $n_c$ = 30 in this work. Experiments were run as 5-fold cross-validation before evaluating the resulting models on the 38-patient test set. 

\begin{table*}[t]
\centering
\caption{Results on the test set of 38 patients. nnUNet fully-supervised ceiling, nnUNet fully-supervised baseline, semi-supervised pseudo-label baseline and the proposed method in the format average ± std. The best values are highlighted in bold.}\label{tab:results2}
\begin{tabularx}{\linewidth}{@{}P{0.3\textwidth}p{0.09\textwidth}*{6}{C}@{}}
\hline
\textbf{Method} & \textbf{WT DSC} & \textbf{TC DSC} &  \textbf{WT HD-95} & \textbf{TC HD-95} & \textbf{WT mIoU} & \textbf{TC mIoU}  \\ 
\hline
Fully supervised ceiling (331 labelled) & 0.904 ±0.006 & 0.869 ±0.004 & 6.099 ±0.496 & 4.128 ±0.371 & 0.836 ±0.006 & 0.802 ±0.004 \\
\hline
Fully supervised baseline (5 labelled) & 0.794 ±0.006 & 0.624 ±0.045 & 14.588 ±1.386 & 17.671 ±2.559 & 0.682 ±0.011 & 0.529 ±0.050 \\
\hline
Semi-supervised baseline (5 labelled, 259 unlabelled) & 0.799 ±0.025 & 0.636 ±0.070 & \textbf{14.244 ±1.873} & 15.164 ±4.782 & 0.686 ±0.033 & 0.548 ±0.076 \\
\hline
Our method (5 labelled, 259 unlabelled) & \textbf{0.824 ±0.023} & \textbf{0.707 ±0.027} & 14.295 ±3.094 & \textbf{14.735 ±3.385} & \textbf{0.722 ±0.025} & \textbf{0.621 ±0.024} \\
\hline
\end{tabularx}
\end{table*}

\begin{figure}[t]
    \begin{subfigure}{0.13\textwidth}
        \includegraphics[width=\linewidth]{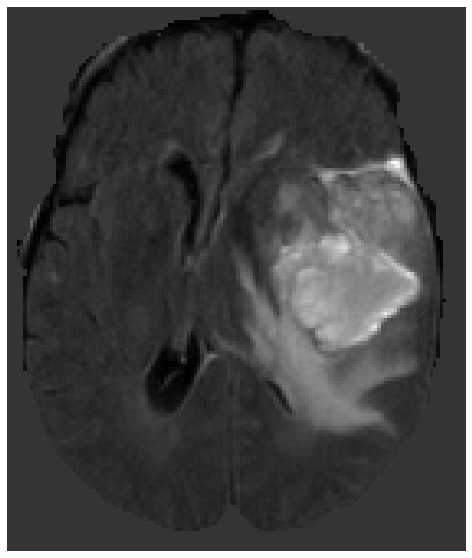}
        \caption{}
        \label{fig:region_growing_a}
    \end{subfigure}%
    \begin{subfigure}{0.11\textwidth}
        \includegraphics[width=\linewidth]{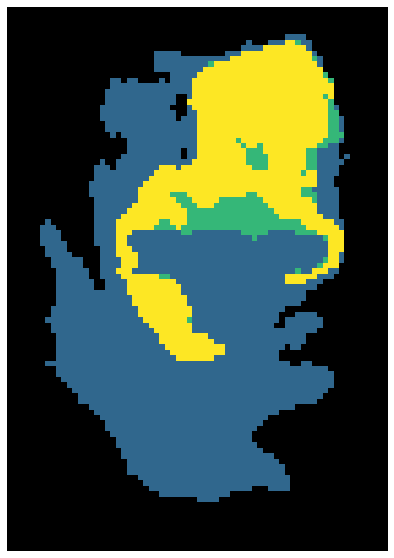}
        \caption{}
        \label{fig:region_growing_b}
    \end{subfigure}%
    \begin{subfigure}{0.11\textwidth}
        \includegraphics[width=\linewidth]{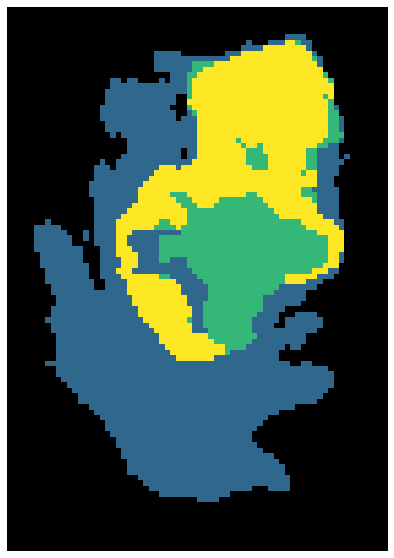}
        \caption{}
        \label{fig:region_growing_c}
    \end{subfigure}%
    \begin{subfigure}{0.11\textwidth}
        \includegraphics[width=\linewidth]{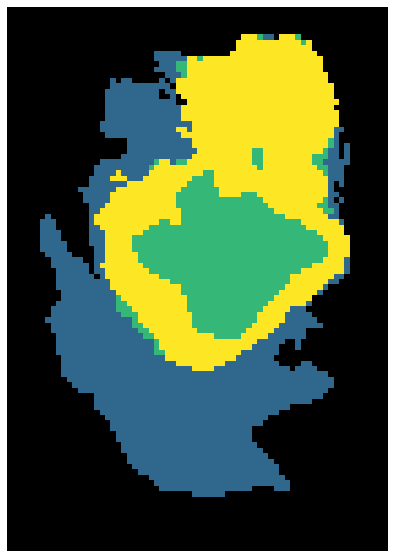}
        \caption{}
        \label{fig:region_growing_d}
    \end{subfigure}
    \caption{Pseudo-label refinement for patient 224 after 200 epochs training (a) FLAIR image (b) network-inferred pseudo-label (c) pseudo-label post 3-D superpixel refinement and (d) GT label}
    \label{fig:region_growing}
\end{figure}

\section{RESULTS \& DISCUSSION}
\label{sec:page}

\autoref{tab:results2} shows the results for the 38-patient hold-out test set. To evaluate the performance of the proposed method, it was compared with a fully-supervised method for all the training data (331 patients: 264 training, 67 validation) which gives a performance ceiling, a fully-supervised baseline (72 patients: 5 training, 67 validation) and a semi-supervised pseudo-labelling baseline method inspired by Lee (2013) \cite{Dong-HyunLee2013} (331 patients: 264 training (5 labelled, 259 unlabelled, 67 validation). The proposed method used the same setup as the pseudo-label-based semi-supervised method, with the addition of the superpixel region refinement algorithm applied to the pseudo-labels after every update step. We report three different metrics to evaluate the segmentation quality: Dice (DSC), 95\% Hausdorff Distance (HD-95) and the mean Intersection-over-Union (mIoU). The best results in \autoref{tab:results2} are highlighted in bold. In general, \autoref{tab:results2} shows that our proposed method achieves a consistent improvement in performance across the three metrics compared to the baseline methods. We report DSC=0.824 and DSC=0.707, HD-95=14.295 and HD-95=14.735, mIoU=0.722 and mIoU=0.621 for the WT and TC regions respectively, when trained on only 5 labelled patients and 259 unlabelled patients. \autoref{fig:region_growing} shows the refinement of the pseudo-label for patient 224 in the training set.  \autoref{fig:region_growing} (c) clearly shows that the superpixel refinement has improved the inferred pseudo-label shown in \autoref{fig:region_growing} (b) and is now closer to the ground truth label shown in \autoref{fig:region_growing} (d).

These results demonstrate the ability of the proposed method, in particular the use of information from multiple modalities, in merging visually non-homogeneous superpixels which belong to the same semantic region, thus overcoming a problem known as ``semantic gap". Further, this method is able to deal with the irregular shapes typical of pathology, such as brain tumours, something superpixels are well suited for. In general, we show that refined pseudo labels result in a better trained model. Given the promising initial results reported in \autoref{tab:results2}, we plan, in future work, to vary the number of labelled patients as well as apply our method to some other medical imaging datasets to determine its robustness.

\section{CONCLUSIONS}
\label{sec:conclusions}
This work investigated the problem of inaccurate pseudo-labels in semi-supervised learning. To overcome this, a simple 3D superpixel region growing algorithm has been developed which refines the pseudo-labels as an update step during training. The results show that, by utilising the features and edges predefined by superpixels, the pseudo-labels, and consequently training, can be improved as shown by the stated performance boost. Although this method has been demonstrated for the task of brain tumour segmentation in MRI, this proof of concept demonstrates the robustness of the proposed method to objects with varying topology and low contrast, which is of use to a wide variety of problems in the medical imaging domain. Furthermore, this method may be used to determine the level of supervision required in future biomedical projects, thus alleviating the annotator burden. 

\section{COMPLIANCE WITH ETHICAL STANDARDS}
This research study was conducted retrospectively using
human subject data made available in open access by the
BraTS 2020 challenge. Ethical approval was not required as
confirmed by the license attached with the open access data.


\bibliographystyle{IEEEbib}
\small
\bibliography{ISBI}

\end{document}